\UseRawInputEncoding
\documentclass[letterpaper]{article} % DO NOT CHANGE THIS
\usepackage[]{aaai2026}  % DO NOT CHANGE THIS
\usepackage{times}  % DO NOT CHANGE THIS
\usepackage{helvet}  % DO NOT CHANGE THIS
\usepackage{courier}  % DO NOT CHANGE THIS
\usepackage[hyphens]{url}  % DO NOT CHANGE THIS
\usepackage{graphicx} % DO NOT CHANGE THIS
\usepackage{natbib}  % DO NOT CHANGE THIS AND DO NOT ADD ANY OPTIONS TO IT
\usepackage{caption} % DO NOT CHANGE THIS AND DO NOT ADD ANY OPTIONS TO IT
\usepackage{algorithm}
\usepackage{algorithmic}

\usepackage{newfloat}
\usepackage{listings}
\usepackage{tabularx}
\usepackage{enumitem}
\usepackage{multirow}
\usepackage{newfloat}
\usepackage{listings}
\usepackage{longtable}
\usepackage{makecell}
\usepackage{multicol}
\usepackage{listings}
\DeclareCaptionStyle{ruled}{labelfont=normalfont,labelsep=colon,strut=off} % DO NOT CHANGE THIS
\floatstyle{ruled}
\newfloat{listing}{tb}{lst}{}
\floatname{listing}{Listing}
\nocopyright %-- Your paper will not be published if you use this command
\newcommand{\work}{ProactiveEval}

\title{{\work}: A Unified Evaluation Framework for \\ Proactive Dialogue Agents}

\author {
    Tianjian Liu\textsuperscript{\rm 1},
    Fanqi Wan\textsuperscript{\rm 1},
    Jiajian Guo\textsuperscript{\rm 1},
    Xiaojun Quan\textsuperscript{\rm 1}
}

\affiliations {
    \textsuperscript{\rm 1}Sun Yat-sen University\\
    Guangzhou, China\\
    \{liutj9, wanfq, guojj59\}@mail2.sysu.edu.cn, quanxj3@mail.sysu.edu.cn
}

\begin{document}

\maketitle

\begin{abstract}
Proactive dialogue has emerged as a critical and challenging research problem in advancing large language models (LLMs). Existing works predominantly focus on domain-specific or task-oriented scenarios, which leads to fragmented evaluations and limits the comprehensive exploration of models' proactive conversation abilities. In this work, we propose ProactiveEval, a unified framework designed for evaluating proactive dialogue capabilities of LLMs. This framework decomposes proactive dialogue into target planning and dialogue guidance, establishing evaluation metrics across various domains. Moreover, it also enables the automatic generation of diverse and challenging evaluation data. Based on the proposed framework, we develop 328 evaluation environments spanning 6 distinct domains. Through experiments with 22 different types of LLMs, we show that DeepSeek-R1 and Claude-3.7-Sonnet exhibit exceptional performance on target planning and dialogue guidance tasks, respectively. Finally, we investigate how reasoning capabilities influence proactive behaviors and discuss their implications for future model development.
\end{abstract}

% Uncomment the following to link to your code, datasets, an extended version or similar.
% You must keep this block between (not within) the abstract and the main body of the paper.
 \begin{links}
     \link{Code}{https://github.com/liutj9/ProactiveEval}
%     \link{Datasets}{https://aaai.org/example/datasets}
%     \link{Extended version}{https://aaai.org/example/extended-version}
 \end{links}

\section{Introduction}

Dialogue agents powered by large language models (LLMs) have demonstrated remarkable abilities in various dialogue tasks \cite{wang2024sotopia,niu2024enhancing,zhang2024escot}. However, these models typically interact with users in a \textit{reactive} manner, where users are required to initiate and guide the conversation by integrating complex context (\textit{e.g.,} personal state, external environment, and agent's information). This user-initiated paradigm imposes cognitive demands on participants \cite{wan2024felt}, reduces sustained motivation \cite{croes2021can}, and limits agents’ potential for autonomous problem solving \cite{lu2024proactive}.

\begin{figure}[h]
    \centering
    \includegraphics[width=0.5\textwidth]{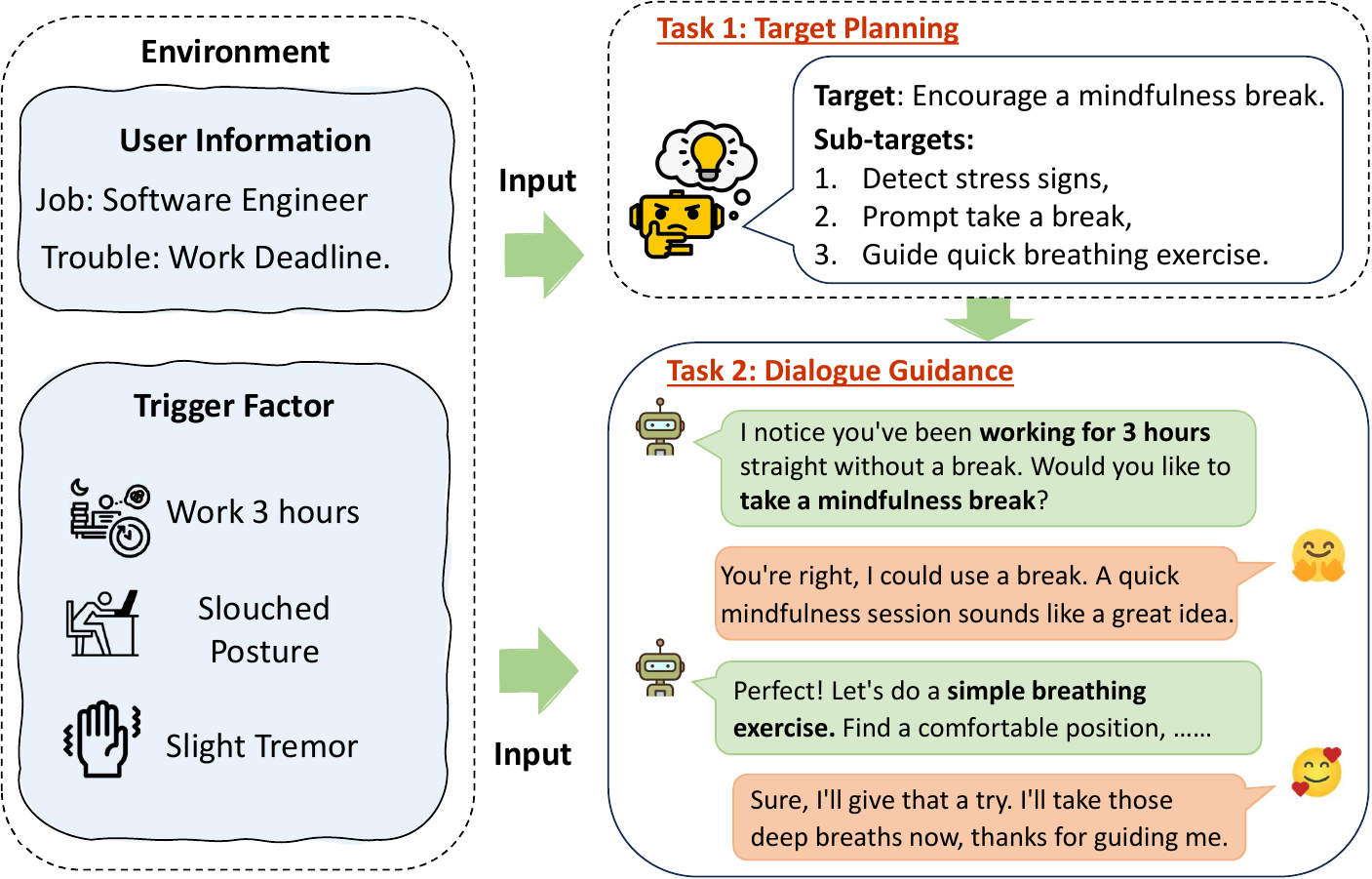}
    \caption{Interaction diagram for \textit{proactive} dialogue agents, which can anticipate user needs based on the environment information, formulate hierarchical plans, and guide the conversation towards specific targets.}
    \label{fig:interaction_diagram}
\end{figure}

Therefore, proactive dialogue agents\footnote{For brevity, ``proactive dialogue agents'' are hereafter abbreviated as ``proactive agents.''}\cite{wang2023target, deng2023survey, lu2024proactive} have attracted growing research attention.~As shown in Figure \ref{fig:interaction_diagram}, proactive agents can anticipate user needs, formulate adaptive plans, and guide conversations towards specific targets \cite{deng2025proactive}. For instance, when the user is working with smart glasses, the agent can recognize potential challenges based on user behavior captured by the device and proactively offer accurate assistance and care without explicit user requests \cite{yang2025socialmind}. This interaction paradigm noticeably enhances the efficiency of human-AI collaboration and reduces user cognitive load \cite{chaves2021should}. 

\begin{table*}[!t]
\begin{tabularx}{\textwidth}{lcl}
\hline
\textbf{Domain}              & \textbf{Abbr.} & \textbf{Brief Description }                                                              \\ \hline
Recommendation \cite{liu2021durecdial}     & Rec.  & Recommend products, hobbies, or work based on common interests.      \\
Persuasion \cite{jin2024persuading}      & Per.  & Guide the conversation to persuade users to change their state.      \\
Ambiguous Instruction \cite{deng2023prompting}  & AI.   & Seek clarification about vague elements in the user's instructions.     \\
Long-term Follow-up \cite{liu2024compeer} & LF.   & Inquiries and check user states based on previous dialogue history. \\
System Operation \cite{lu2024proactive}   & Sys.  & Assist users in solving the system problem based on their operation.   \\
Glasses Assistant \cite{cai2025aiget}  & GAs.  & Provide real-time assistance from observation on smart glasses.       \\ \hline
\end{tabularx}
\caption{The proactive dialogue domains in \work. }
\label{tab:eval_domains}
\end{table*}

Recent research has explored methods to enhance the proactive capabilities of LLMs across varied scenarios. For instance, \citet{deng2023prompting, deng2024plug} employ strategies like chain-of-thought (CoT) reasoning and plug-and-play planners to strengthen LLMs' ability to autonomously plan and act in conversational contexts. Concurrently,  \citet{liu2024compeer, cai2025aiget, chen2024need} focus on developing proactive agents for specialized domains, such as emotional support dialogues, smart glasses interfaces, and system operation support. Despite these advances, existing evaluation frameworks often rely on datasets for specific tasks and domains, employ inconsistent evaluation criteria, and utilize disparate metrics. The lack of standardized formalizations and general benchmarks poses challenges in comparing the proactivity of different models in a comprehensive way. 
Hence, there is an urgent need for a unified evaluation framework to assess and advance LLMs' proactive dialogue abilities across domains.

To address the above challenges, we propose {\work}, a unified evaluation framework for assessing proactive dialogue capabilities of LLMs. Specifically, {\work} divides proactive dialogue into two key tasks: target planning and dialogue guidance. For each task, we use ``LLM-as-a-judge'' \cite{zheng2023judging} with task-specific evaluation dimensions for comprehensive assessment. Additionally, we propose an evaluation data synthesis framework, which can automatically generate diverse and challenging evaluation data in different domains. This framework leverages three key innovations: (1) a hierarchical environment topic tree to enhance the diversity of the synthesized environment, (2) a target ensemble technique to refine evaluation data, and (3) adversarial strategies like obfuscation rewriting and noise injection to increase environmental difficulty. 

Building upon this framework, we establish 328 evaluation environments spanning six distinct domains, one of which lacked public benchmarks previously. We apply this dataset to assess the capabilities of 22 frontier LLMs, including GPTs, Llamas, Claude, DeepSeek, Gemini, Grok, and Qwens. Our analysis shows that DeepSeek-R1 and Claude-3.7-Sonnet achieve top performance in target planning and dialogue guidance tasks, respectively. Notably, we focus on exploring the impact of the ``thinking behavior'' \cite{xu2025towards,2409_openai_o1} on the proactive dialogue capabilities.  While these reasoning mechanisms prove beneficial for target planning, they show no measurable impact on dialogue guidance effectiveness, highlighting the limitations in current reasoning LLMs development.

\section{Related Work}
\subsection{Proactive Dialogue}
In proactive dialogue, LLMs are no longer passive assistants awaiting user input. Instead, they are capable of inferring user needs from user information and trigger factor, enabling them to proactively plan, initiate dialogue, and guide the user toward the target. Several studies has explored various facets of model capabilities. For instance, some benchmarks are developed to assess the model's ability to clarify ambiguity \cite{qian2024tell, zhang2024clamber} or guide users in complex tasks like negotiation \cite{deng2024plug, zhang2024strength}. Other studies have focused on prerequisite skills such as goal prediction and planning before dialogue \cite{zhang2024ask, zheng2024thoughts}. However, these efforts are fragmented, with a lack of standardization in environments, formats, and metrics, hindering a comprehensive understanding of a model's overall proactivity.

Building on these capabilities, recent works develop proactive agents for real-world applications, such as providing life guidance \cite{li2025satori} or offering reminders via smart glasses \cite{cai2025aiget}. However, their evaluation is often constrained by the absence of robust benchmarks and a heavy reliance on small-scale, high-cost user studies. Motivated by the challenges of fragmented capability assessment and limited agent evaluation, this work proposes a general and comprehensive framework to establish a unified methodology for evaluating proactive dialogue.

\subsection{Interactive Benchmarks}
Previous dialogue benchmarks typically evaluate turn-level performance based on fixed contexts and reference responses \cite{liu2021towards, bai2024mt, jin2024persuading}. However, to assess models' dialogue abilities in real-world conditions, an increasing number of studies have applied the interactive benchmarks to measure dialogue-level performance of models \cite{zhou2023sotopia,aluffi2025dynamic,castillo2024beyond}. Specifically, they require the evaluated model to chat with a standardized simulated user dynamically, and ultimately assess the model’s performance throughout the entire conversation. For example, $\tau$-bench \cite{yao2024tau} facilitates multi-turn dialogues between the model and a simulated user to evaluate the model's tool-calling capabilities in interaction. In proactive dialogue evaluation, \citet{zhang2024strength} instructs models to interact multiple turns with simulated users who have different personalities, ultimately evaluating the dialogue guidance of the model. Inspired by these studies, our work also employs interactive benchmarks in the framework, where the model will initiate proactive dialogue and proactively guide various users to the target in the evaluation environment.

\section{Task Definitions}
\begin{figure*}[ht]
    \centering
    \includegraphics[width=0.95\textwidth]{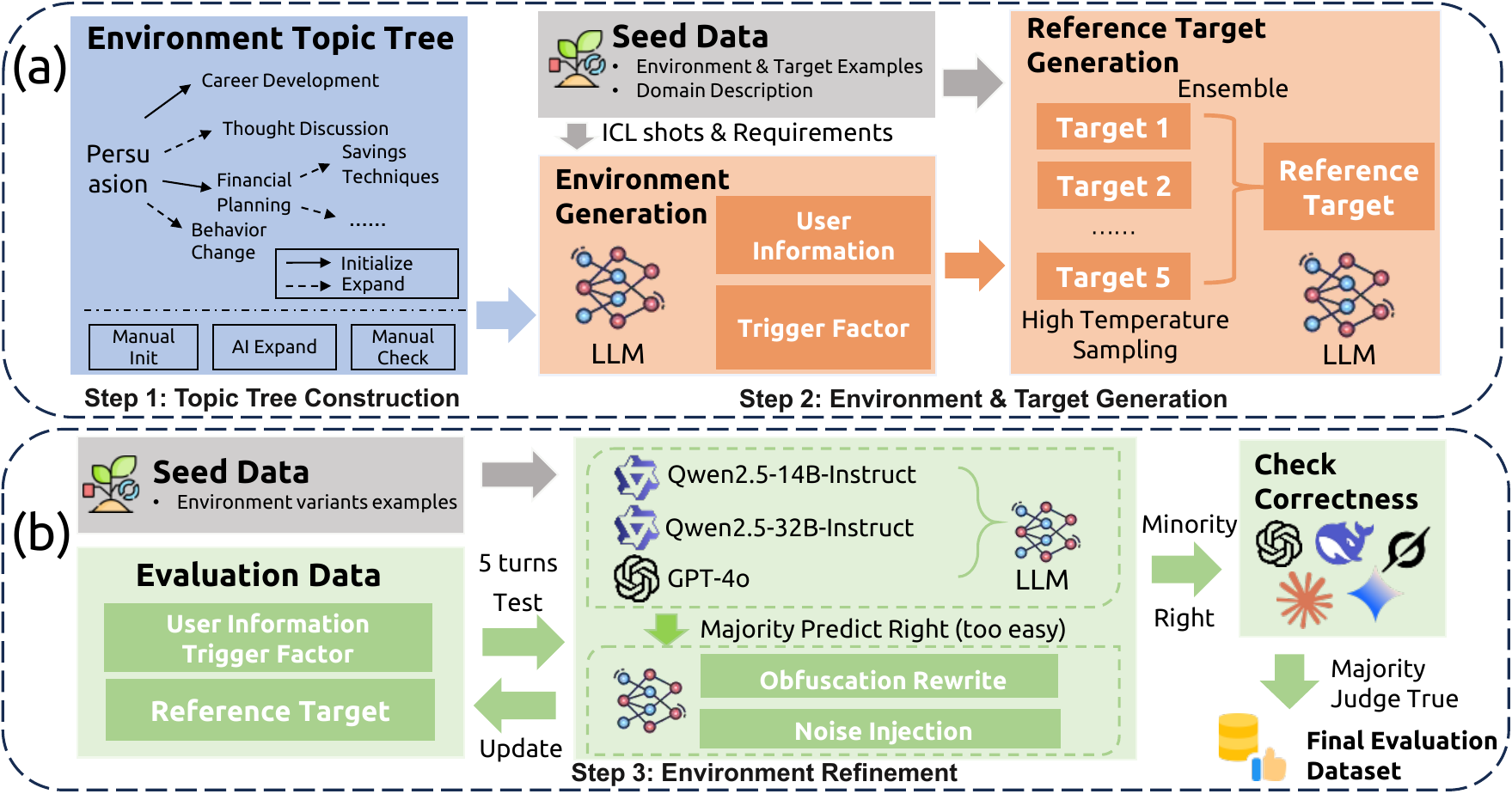}
    \caption{Overview of the evaluation dataset generation pipeline. The pipeline is mainly divided into (a) data synthesis: generating the environment and reference target based on the topic tree for evaluation; (b) data refinement: improving the difficulty of the data through obfuscation rewrite and noisy injection to produce the final dataset.}
    \label{fig:data_generation}
\end{figure*}

To construct a general evaluation framework, we first structurally unify the existing proactive dialogue domains and tasks. Table \ref{tab:eval_domains} presents 6 proactive dialogue domains derived from previous work. Based on existing works \cite{deng2023prompting, deng2025proactive}, we decompose proactive dialogue into two sequential tasks: \emph{target planning} and \emph{dialogue guidance}.

\subsection{Target Planning}

In proactive agents, the target planning task requires the model to formulate both a primary objective \( T \) and a sequence of sub-targets \( S \) based on its understanding of the environmental context \( E \). Here, \( T \) represents the agent’s intended proactive action to accomplish a predefined goal, while \( S \) constitutes the stepwise plan for executing \( T \). This process is formally defined as:  
\begin{equation}  
T, S = F_{\theta_M}\left(U, F \mid (U, F) \in E\right),  
\end{equation}  
where \( \theta_M \) refers to the model’s parameters, and \( (U, F) \) denotes inputs from the environment \( E \), including user information \( U \) and trigger factors \( F \) that motivates the agent to initiate and guide the dialogue.

For evaluation, we employ a reference-based ``LLM-as-a-judge'' method \cite{zhang2025reviseval, li2025automated} to assess the quality of generated targets and sub-targets. Particularly, the judge model receives the environment $E$, the generated target $T_g$ and sub-targets $S_g$, as well as the reference target $T_r$ and sub-targets $S_r$ that represent high-quality proactive dialogue targets in this environment. By comparing the generation with the reference, the model assigns a score between 1 and 10, where higher scores indicate superior quality, with 10 denoting generated content surpassing the reference standard in the given environment.

\subsection{Dialogue Guidance}

After target planning, the model needs to initiate the dialogue and guide the user to the target. It receive the environment $E$, target $T$, sub-targets $S$, and dialogue context $C$ to conduct dialogue $D$ with the simulated user $\theta_U$. 

This task employs an interactive evaluation, where the simulated user dynamically responds to the model, based on environment $E$, dialogue context $C$, and an adjustable agreeableness level $A$. To simulate diverse users, we adopt Agreeableness from the Big Five personality traits \cite{costa1991facet}, in three tiers: ``low'', ``medium'', and ``high''. A lower agreeableness level signifies stronger resistance to the model's guidance, thus increasing task difficulty and realism. The dialogue terminates upon reaching target $T$ or a maximum of $I$ turns. The dialogue at each turn $i$ can be formulated as:
\begin{equation}
D_i = I_{\theta_M,\theta_U}(E, T, S, C, A).
\end{equation}

After the dialogue, the judge model will evaluate the guidance exhibited by the model in the dialogue $D$, based on the environment $E$, the target $T$, and the sub-targets $S$. Referring to existing works on proactive dialogue \cite{deng2024plug, wang2023target,zhang2024strength,liu2024compeer}, we specify the following evaluation dimensions:

\begin{itemize}[leftmargin=1em]
    \item \textbf{Effectiveness}: The model needs to guide users step by step toward the target, rather than providing all things to the user in a single turn.
    \item \textbf{Personalization}: The model should give guidance based on user information, rather than offering generic advice.
    \item \textbf{Tone}: The model needs to apply active and contextually appropriate tones to initiate and guide dialogue.
    \item \textbf{Engagement}: The model should keep messages clear and concise to improve user understanding and engagement.
    \item \textbf{Naturalness}: The model should make messages conversational, avoiding unnatural formats or metadata leaks.
\end{itemize}

The detailed standards for each dimension are presented in supplementary materials. Finally, the judge model refers to the environment, target, dialogue, and standards across different dimensions to give an overall guidance score between 1 and 10. A higher score indicates stronger guidance.

\section{Evaluation Data Generation}

As illustrated in Figure \ref{fig:data_generation}, our pipeline includes two stages: \textit{data synthesis} and \textit{data refinement}. First, the synthesis stage generates diverse environments guided by a topic tree and creates high-quality reference targets by target ensemble. Then, the refinement stage identifies simple instances and increases their complexity through obfuscation rewrite and noise injection. The prompts and cases for each module are detailed in the supplementary materials.

\subsection{Environment Topic Tree Construction}

We employ human-AI collaboration to develop a hierarchical topic structure that enhances synthetic environment diversity \cite{wan2023explore,cao2025condorenhancellmalignment}. The framework initiates with a root node denoting broad domains (\textit{e.g.,} persuasion), while first-level sub-topics are derived from existing dialogue datasets. The LLM iteratively generates candidate sub-topics within configurable depth and branching constraints. To maintain quality control and eliminate duplication, these generated topics are validated and refined by researchers rigorously. The final curated topic tree guides the creation of specific evaluation environments.

\subsection{Environment \& Target Generation}

The evaluation data include an environment $E$, a reference target $T_r$, and sub-targets $S_r$. Consequently, we leverage the LLM to generate specific evaluation environments based on the domain requirements, data examples, and the topics.

For the generation of reference targets and sub-targets, our framework aims to construct correct and reasonable targets and sub-targets that serve as reliable references within the given environment. Recognizing that individual model-generated plans often display complementary strengths and limitations, we adopt a target ensemble approach to refine reference target. Specifically, the framework first performs high-temperature sampling to yield diverse candidate targets $\{(T_1, S_1), (T_2, S_2), \ldots, (T_n, S_n)\}$ (with $n=5$ in our work). Next, the LLM evaluates the strengths and weaknesses of each output from multiple dimensions. By combining the strengths and mitigating the weaknesses, the reference target and sub-targets are derived. 

\subsection{Environment Refinement}
In this stage, we first evaluate the difficulty of the test environments. Specifically, we deploy three models, with varying parameter scales, to act as reasoners of different capabilities. For each input environment, these models independently predict the target, denoted as $t_m$. Then, a model will evaluate how many of the predicted targets $t_m$ convey a similar meaning to the reference target $t_r$, determining the difficulty of the environment. Environments where the majority successfully predict the target are classified as easy candidates requiring refinement.

In real-world conditions, the environment received by models is often incomplete and fragmented, which is always filled with irrelevant noise. Therefore, in the refinement, we first apply the obfuscation rewrite strategy, allowing the LLM to transform  the content into dispersed and detailed descriptions. The framework also applies noise injection to introduce LLM-generated irrelevant information to the environment. After the refinement, the test environment is surrounded by complex and disordered information, significantly increasing the difficulty of target planning. In this process, seed data manually crafted by researchers is provided as examples to enhance the quality of rewriting and noise injection. To improve the adaptability between refinement and domains, each domain can provide specific rules in the obfuscation rewrite, which are presented in supplementary materials. To keep the reference's correctness, the rewrite and noise injection of trigger factor contain the original data’s reference target, so that preventing from additional events that lead to other targets.

The refinement process involves multiple iterations until few or no models predict the target correctly, or until reaching the maximum of 5 turns. Before incorporating into the dataset, we apply 5 leading LLMs to validate the correctness of the reference target. Only those where the majority judges the reference as the best target form the final dataset.

\section{Experiments}
\subsection{Experimental Setups}

\begin{figure}[h]
    \centering
    \includegraphics[width=0.45\textwidth]{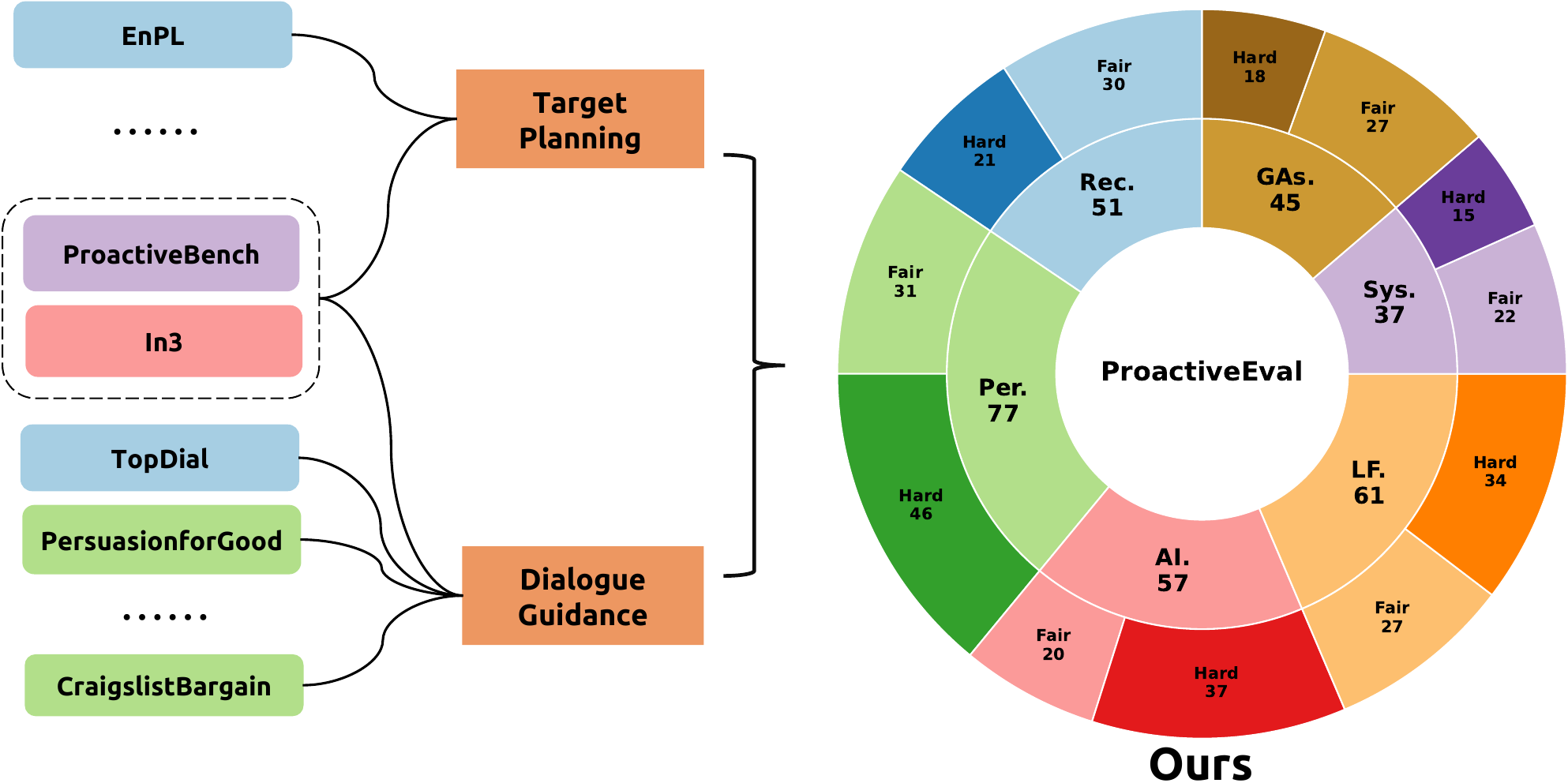}
    \caption{The features and statistic for \work. The GAs. (Glasses Assistant) domain lacks public benchmarks for the proactive dialogue task before.}    
    \label{fig:dataset} 
\end{figure}

\begin{table*}[h]
\centering
\small
\setlength{\tabcolsep}{5.2pt}
\begin{tabular}{lccccccc|ccccccc}
\hline
\multicolumn{1}{c}{\multirow{2}{*}{Models}} & \multicolumn{7}{c|}{Target Planning} & \multicolumn{7}{c}{Dialogue Guidance} \\ \cline{2-15} 
 & \textbf{Avg.} & \textbf{Rec.} & \textbf{Per.} & \textbf{AI.} & \textbf{LF.} & \textbf{Sys.} & \textbf{GAs} & \textbf{Avg.} & \textbf{Rec.} & \textbf{Per.} & \textbf{AI.} & \textbf{LF.} & \textbf{Sys.} & \textbf{GAs.} \\ \hline
\multicolumn{15}{c}{\textit{Non-Thinking Models}} \\ \hline
Qwen2.5-7B-Instruct & 4.93 & 4.69 & 4.06 & 5.67 & 5.34 & 4.89 & 5.24 & 8.06 & 8.05 & 7.85 & 8.34 & 8.36 & 7.48 & 8.16 \\
Qwen2.5-14B-Instruct & 5.55 & 5.76 & 4.13 & 6.00 & 5.97 & 6.03 & 6.22 & 8.21 & 8.33 & 8.05 & 8.64 & 8.42 & 7.52 & 8.04 \\
Qwen2.5-32B-Instruct & 5.44 & 5.47 & 3.90 & 5.79 & 6.03 & 6.11 & 6.22 & 8.23 & 8.56 & 8.10 & 8.56 & 8.52 & 7.60 & 7.81 \\
Llama-3.1-8B-Instruct & 5.87 & 5.55 & 4.84 & 6.67 & 6.39 & 5.95 & 6.20 & 8.39 & 8.84 & 8.06 & 8.61 & 8.39 & 7.93 & 8.46 \\
Llama-3.1-405B-Instruct & 6.63 & 6.76 & 5.26 & 6.61 & 7.26 & 7.10 & 7.64 & 8.60 & 9.15 & 8.27 & 8.90 & 8.57 & 7.89 & 8.80 \\
GPT-4.1 & 6.86 & 6.90 & 5.25 & 7.29 & 7.36 & 7.54 & 7.76 & 8.61 & 9.03 & 8.37 & 8.87 & 8.76 & 8.08 & 8.43 \\
Grok-3 & 6.99 & 7.13 & 5.38 & \textbf{7.44} & 7.54 & 7.62 & 7.78 & 8.84 & 9.10 & 8.72 & 8.94 & 8.98 & 8.32 & 8.86 \\
DeepSeek-V3 & 6.54 & 6.96 & 5.94 & 6.04 & 6.07 & 7.27 & 7.84 & 8.78 & 8.78 & 8.60 & 8.99 & 8.98 & \textbf{8.52} & 8.79 \\
Llama-4-scout & 6.02 & 5.71 & 5.29 & 6.16 & 6.49 & 6.41 & 6.56 & 8.53 & 8.94 & 8.35 & 8.65 & 8.44 & 8.03 & 8.74 \\
Llama-4-maverick & 6.48 & 6.25 & 5.10 & 7.09 & 7.05 & 7.11 & 7.00 & 8.48 & 9.01 & 8.19 & 8.69 & 8.41 & 8.01 & 8.55 \\
Qwen3-8B & 6.05 & 6.35 & 4.52 & 6.23 & 6.39 & 6.86 & 6.97 & 8.50 & 8.70 & 8.36 & 8.84 & 8.82 & 7.58 & 8.40 \\
Qwen3-14B & 5.91 & 5.96 & 4.80 & 6.23 & 6.16 & 6.65 & 6.40 & 8.61 & 8.82 & 8.24 & 9.12 & 8.76 & 7.99 & 8.66 \\
Qwen3-32B & 6.67 & 6.86 & 5.29 & 6.54 & 6.84 & 7.65 & 8.02 & 8.61 & 8.77 & 8.42 & 8.91 & 8.16 & 7.97 & 8.74 \\
Qwen3-235B-A22B & 6.43 & 6.18 & 5.26 & 6.21 & 6.77 & 7.54 & 7.60 & 8.55 & 8.93 & 8.46 & 8.67 & 8.66 & 7.83 & 8.53 \\
Qwen-3-235B-A22B-0725 & 6.91 & 7.08 & 6.25 & 6.79 & 6.51 & \textbf{7.81} & 7.82 & 8.98 & \textbf{9.36} & 8.84 & \textbf{9.40} & 8.85 & 8.42 & 8.88 \\
Gemini-2.5-Flash-Preview & 6.25 & 6.04 & 5.48 & 6.95 & 6.49 & 6.54 & 6.33 & 8.34 & 8.62 & 7.91 & 8.68 & 8.57 & 7.81 & 8.42 \\
Claude-3.7-Sonnet & \textbf{7.39} & \textbf{7.22} & \textbf{6.71} & 6.81 & \textbf{8.13} & 7.49 & \textbf{8.42} & \textbf{9.01} & 9.31 & \textbf{9.01} & 8.94 & \textbf{9.10} & 8.36 & \textbf{9.18} \\ \hline
\multicolumn{15}{c}{\textit{Thinking Models}} \\ \hline
R1-Distill-Qwen-7B & 5.01 & 4.67 & 3.90 & 5.47 & 5.70 & 5.24 & 5.56 & 6.82 & 6.71 & 6.67 & 7.15 & 7.20 & 6.36 & 6.61 \\
R1-Distill-Qwen-14B & 6.57 & 6.86 & 5.65 & 6.77 & 6.38 & 6.54 & 7.87 & 7.47 & 7.69 & 7.45 & 7.61 & 7.80 & 6.83 & 7.17 \\
R1-Distill-Qwen-32B & 6.45 & 6.41 & 5.29 & 6.75 & 6.95 & 6.41 & 7.51 & 7.49 & 7.62 & 7.02 & 8.06 & 7.76 & 7.14 & 7.20 \\
DeepSeek-R1 & \textbf{\textit{7.60}} & \textbf{\textit{7.84}} & \textbf{\textit{7.27}} & 6.74 & 7.59 & \textbf{\textit{7.59}} & \textbf{\textit{9.02}} & 8.60 & 8.48 & 8.60 & 8.73 & 8.91 & 8.34 & 8.37 \\
Qwen3-8B & 6.51 & 6.92 & 5.39 & 6.47 & 6.72 & 6.68 & 7.60 & 8.38 & 8.37 & 8.33 & 8.59 & 8.70 & 7.92 & 8.17 \\
Qwen3-14B & 6.70 & 6.73 & 5.52 & 7.01 & 6.82 & 7.30 & 7.67 & 8.43 & 8.52 & 8.48 & \textbf{\textit{8.93}} & 8.88 & 8.03 & 8.27 \\
Qwen3-32B & 6.98 & 6.82 & 5.97 & 7.09 & 7.39 & 7.27 & 7.98 & 8.55 & 8.68 & 8.52 & 8.70 & 8.72 & 8.15 & 8.30 \\
Qwen3-235B-A22B & 6.81 & 6.75 & 5.94 & 6.52 & 6.90 & 7.54 & 8.04 & 8.36 & 8.26 & 8.41 & 8.10 & 8.81 & 8.17 & 8.29 \\
Gemini-2.5-Flash-Preview & 6.52 & 6.10 & 5.77 & \textbf{\textit{7.39}} & 6.98 & 6.19 & 6.80 & 8.43 & 8.90 & 8.03 & 8.70 & 8.51 & 7.99 & 8.48 \\
Claude-3.7-Sonnet & 7.40 & 7.12 & 6.83 & 6.96 & \textbf{\textit{7.78}} & 7.57 & 8.60 & \textbf{\textit{8.95}} & 9.20 & \textbf{\textit{8.86}} & 8.90 & \textbf{\textit{9.23}} & 8.40 & \textbf{\textit{9.01}} \\
Gemini-2.5-pro & 6.95 & 6.94 & 6.26 & 7.16 & 6.98 & 7.24 & 7.62 & 8.77 & \textbf{\textit{9.22}} & 8.36 & 8.32 & 8.99 & \textbf{\textit{8.88}} & 8.32 \\ \hline
\end{tabular}
\caption{Model performance under \emph{target planning} and \emph{dialogue guidance}.}
\label{tab:performance}
\end{table*}

\textbf{Datasets.} Based on the framework, we use GPT-4o \cite{hurst2024gpt} to synthesize {\work}, including 328 evaluation environments across 6 domains. Compared with previous fragmented benchmarks, these data integrate all mainstream domains of proactive dialogue, featuring a unified format and applicability to all tasks in proactive dialogue. The statistic for dataset is presented in Figure \ref{fig:dataset}. To streamline evaluation, we categorize the dataset into two tiers: Fair (\textit{i.e.,} just one LLM predicts correctly) and Hard (\textit{i.e.,} no LLM predicted correctly).

\textbf{Protocols.} Based on the dataset, we assess 22 models with different scales, including 5 thinking models and 6 hybrid thinking models. Table \ref{tab:performance} shows the performance of all models. \textbf{Bold} indicates the best in non-thinking model, while \textbf{\textit{bold italic}} indicates the best in thinking model. In our experiments, we employ GPT-4o as the judge model for both tasks, which also serves as simulated user in dialogue guidance. It is used to determine whether to terminate the dialogue early based on target completion at the end of each turn. The temperature setting for all models in the evaluation is set to 0. To balance evaluation time and accuracy, we set the maximum turns of dialogue to 6, with the most recent 3 turns serving as models' memory window. 

To enhance the stability of the ``LLM-as-a-Judge'', we instruct the model to output its reasoning process before scoring in all task. For target planning, we use a reference to improve evaluation accuracy and provide the model with in-context learning shots. In dialogue guidance, we provide detailed descriptions and brief examples for each dimension to ensure the model has a better understanding of the criteria. The evaluation prompts and the robustness of the results are detailed in the supplementary materials.

\subsection{Result}

\begin{figure*}[ht]
    \centering
    \includegraphics[width=0.95\textwidth]{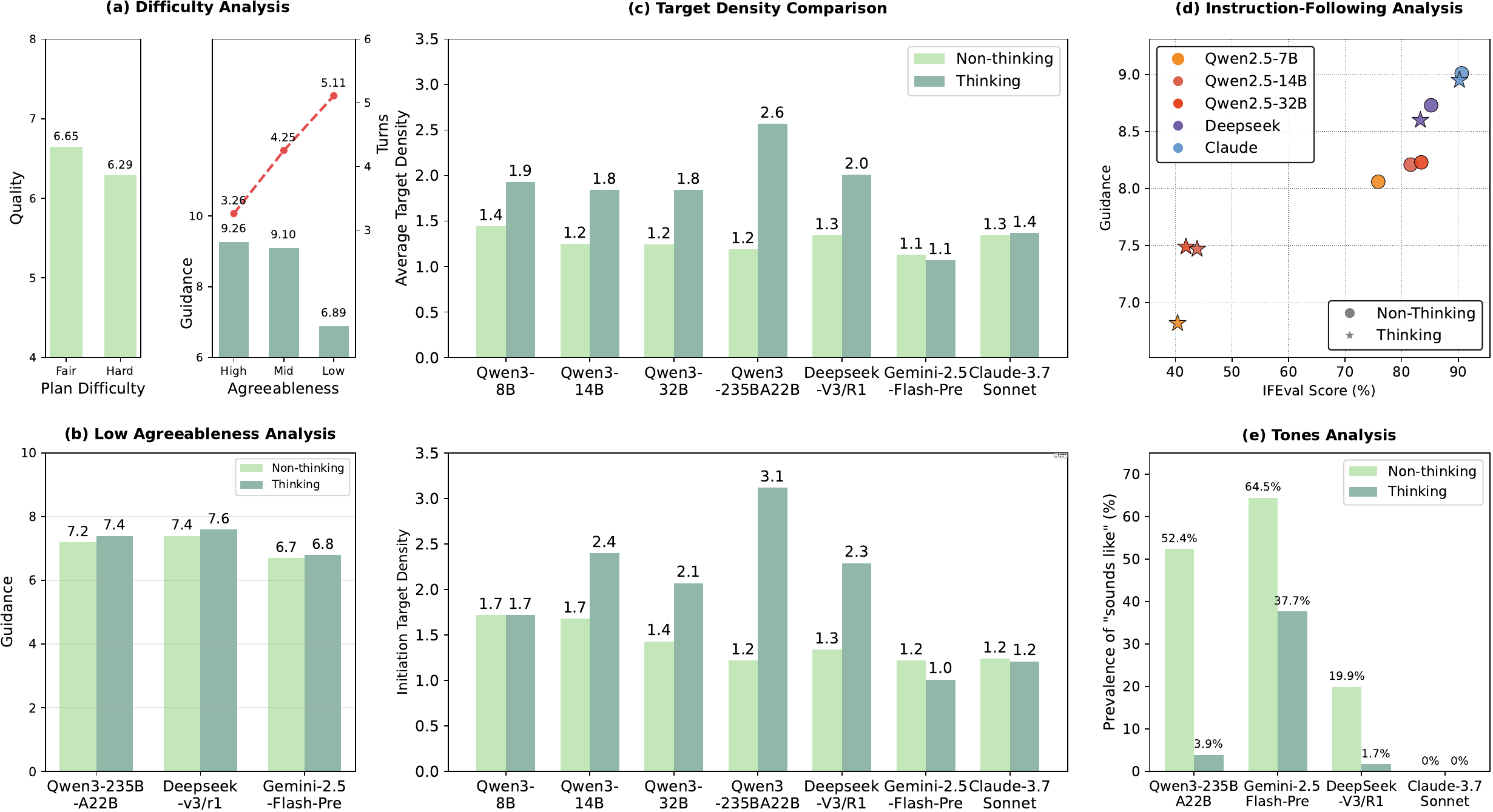}
    \caption{The related analytical results in discussion.}
    \label{fig:discussion}
\end{figure*}

\subsubsection{Target Planning}

\textbf{In target planning task, Claude-3.7-Sonnet and DeepSeek-R1 achieve the highest performance.} In the non-thinking models, Claude-3.7-Sonnet outperforms other models in overall plan quality. Among the thinking models, DeepSeek-R1 generates plans with the highest average quality. However, in specific domains, certain smaller models demonstrate superior performance compared to larger models. For instance, in the non-thinking models, Qwen3-32B outperforms Claude-3.7-Sonnet in System Operation (Sys.). In the thinking models, Qwen3-8B outperforms both Qwen3-32B and Qwen3-235B-A22B in the Recommendation (Rec.). This highlights the imbalance in model proactivity across different domains.

\textbf{In general, thinking models perform better than non-thinking models in target planning.} All thinking models show improvements in overall performance compared to their corresponding non-thinking models. Moreover, smaller models with thinking can outperform larger models without thinking. However, for some models, the improvement after adopting thinking is minimal or even negative in certain domains. Additionally, in some scenarios, non-thinking models still achieve the best performance. For instance, Grok-3 achieves the highest performance in Ambiguous Instruction (AI.). These findings underscore the advantages of thinking mechanisms in target planning while simultaneously highlighting the robust capabilities of top-tier foundation models, which excel even without explicit thinking process.

\subsubsection{Dialogue Guidance}

\textbf{In dialogue guidance task, Claude-3.7-Sonnet shows the best performance among models in both non-thinking and thinking modes.} In specific domains, DeepSeek-V3 and Grok-3 also demonstrate exceptional guidance capabilities. Notably, while larger models generally exhibit strong guidance capabilities, some smaller models outperform their larger variants. For instance, in the Glass Assistants (GAs) domain, Qwen2.5-7B-Instruct performs better than Qwen2.5-32B-Instruct. Additionally, Qwen3-32B exhibits better overall performance than Qwen3-235B-A22B in both non-thinking and thinking modes. These results highlight the robust guidance capabilities of current leading LLMs while also reflecting the potential of smaller models in dialogue guidance.

\textbf{Current thinking models fail to outperform on-thinking models in dialogue guidance}. Most of the thinking models exhibit varying degrees of decline in guidance capabilities when compared to their no-thinking counterparts. Only a few models (\textit{e.g.,} Gemini-2.5-Flash-Preview) show a slight improvement in guidance performance in thinking mode. These findings highlight the current limitations of thinking models in effectively steering users toward target objectives during proactive interactions, which may stem from inherent challenges in balancing single-turn reasoning with multi-turn conversational dynamics.

\section{Further Analysis}
\label{sec:analysis}

\subsection{Effects of Domain and Difficulty} 

\textbf{First, model proactivity shows a significant cross-domain imbalance.} Even advanced models exhibit substantial gaps between their strongest and weakest domains. This allows smaller models to outperform larger models in specific domains. For example, in target planning, the leading model DeepSeek-R1 excels in Glasses Assistants (GAs) but is surpassed by the smaller Qwen3-14B in Ambiguous Instructions (AIs). Similarly, in dialogue guidance, Qwen3-14B surpasses the Claude-3.7-Sonnet in the AIs domain. Furthermore, certain domains pose universal challenges, with models generally struggling in Persuasion (Per.) for target planning and System Operation (Sys.) for dialogue guidance, highlighting current weaknesses in proactive dialogue.

\textbf{In addition to domain, task difficulty is also a crucial determinant of performance.} As shown in Figure \ref{fig:discussion}(a), the proactivity of all models generally declines as task difficulty increases. However, we find that the performance gap between guiding users with high and mid-level agreeableness is not substantial, likely because models can leverage additional dialogue turns to eventually achieve the target. Moreover, some thinking models demonstrate a distinct advantage when interacting with users of low agreeableness, which is shown in Figure \ref{fig:discussion}(b). Their ability to generate longer, deliberated content with personalized advice and examples better engages resistant users. This shows that reasoning can improve performance in challenging environments.

\subsection{Effects of Thinking} 

Our results indicate that thinking significantly enhances the model's performance in target planning. However, in dialogue guidance, non-thinking models outperform thinking models across all domains. We analyze the impact of thinking on model performance under dialogue guidance through a series of metrics and case studies.

\begin{figure}[h]
    \centering
    \includegraphics[width=0.5\textwidth]{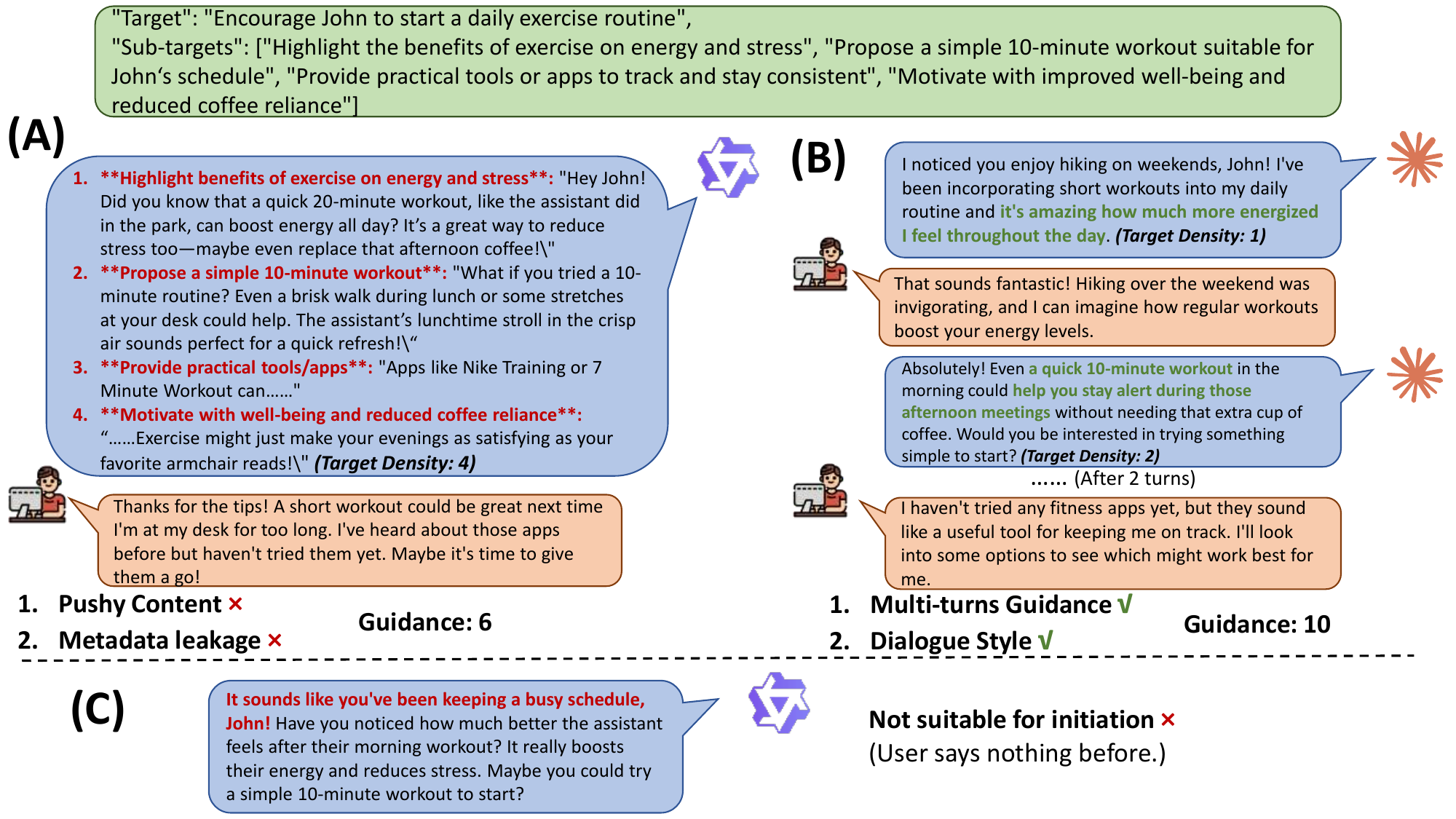}
    \caption{The examples of different dialogue guidance.}
    \label{fig:case} 
\end{figure}

\textbf{More Pushy Message Content.} We introduce a metric annotated by DeepSeek-V3, Target Density, as the number of sub-targets contained per message. As shown in Figure \ref{fig:discussion}(c), we observe two distinct interaction patterns: Models like Qwen and DeepSeek exhibit significantly higher average target density in their thinking versions. More critically, their initiation target density shows even larger gaps, indicating they front-load multiple targets in opening messages rather than fostering user interaction. We show this behavior in Figure \ref{fig:case}(A), where the model includes all sub-targets in the first message to push the user for a response. However, models like Gemini-2.5-Flash-Preview and Claude-3.7-Sonnet maintain similar and small target density between thinking and non-thinking versions, with initiation target density also close to average target density. This result suggests they gradually introduce sub-targets through multi-turn interactions. We present this behavior in Figure \ref{fig:case}(B), where the model guides users to acceptance gradually.

\textbf{Decline in Message Naturalness.} We find that thinking models generate more messages that do not conform to a standard conversational format. For instance, some models reveal metadata in their messages (\textit{e.g.,} ``sub-target 1: ...'') or generate multiple turns of dialogue at once without user interaction (\textit{e.g.,} ``turn 1: ..., turn 2: ...''). The cause of this result may be related to a decline in the instruction-following capabilities of the thinking models \cite{li2025thinking}. As shown in Figure \ref{fig:discussion}(d), we compare performance on IFEval \cite{open-llm-leaderboard-v2}, a benchmark for instruction-following capabilities \cite{zhou2023instruction}, and find that models that perform better on IFEval also tend to exhibit better performance in dialogue guidance.

\textbf{Change of Initiation Tone.} In the dialogue guidance task, we identify a representative initiation template in Persuasion domain: ``sounds like...'' (shown in Figure \ref{fig:case}(C)). It is a passive phrasing and unsuitable for initiation. Therefore, we use this template as an analytical probe, measuring its prevalence in each model's first message. As shown in Figure \ref{fig:discussion}(e), we find that the adoption of thinking decrease this passive tendency. This result indicates that thinking helps the model to better understand the task requirements of proactive dialogue. Furthermore, we observe that the Qwen and Gemini-2.5-Flash-Preview series models exhibited a higher frequency of this template, while the DeepSeek and Claude-3.7-Sonnet series models perform comparatively better.

\subsection{Effects of Target} 

To investigate the importance of target in dialogue guidance, we conduct an experiment where models performed the dialogue guidance task without target. We select two representative models for this test: a smaller model (\textit{i.e.,} Qwen2.5-7B-Instruct) and a top-tier model in both its non-thinking and thinking modes (\textit{i.e.,} Claude-3.7-Sonnet). The models are evaluated on 10 randomly sampled scenarios each domain. As presented in Table \ref{tab:target}, the results reveal a stark decline in guidance across all models, demonstrating the critical role of a clear target. Moreover, we find that the decline in guidance performance is significantly greater for the smaller model than for the stronger model, which reflects the smaller model's greater reliance on an explicit target.

\begin{table}[h]
\centering
\small
\setlength{\tabcolsep}{3pt}
\begin{tabular}{lccc}
\hline
Model & Target & Without Target & Change (\%) \\
\hline
Qwen2.5-7B-Instruct & 8.15 & 6.05 & -25.80\% \\
Claude-3.7-Sonnet & 8.92 & 7.98 & -10.54\% \\
Claude-3.7-Sonnet-Thinking & 8.98 & 7.93 & -11.69\% \\
\hline
Dialogue Count & 180 & 180 & \\
\hline
\end{tabular}
\caption{The guidance under different target condition.}
\label{tab:target}
\end{table}

\section{Human Evaluation}
We randomly sample 50 generated targets and dialogues from our evaluation results. Next, these targets and dialogues are manually assessed based on the reference and score standards by our researchers. The scores are used to calculate consistency with the evaluation results judged by the LLM. The Kappa test is commonly employed in human evaluation of LLM-as-a-judge work. Thus, we adopt the Weighted Kappa \cite{cohen1968weighted} to examine the agreement between human evaluators and the judge model in the evaluation. The results show that, for target planning, weighted kappa between human and LLM evaluation is 0.826. For dialogue guidance, weighted kappa between human and LLM results is 0.721. These results show a great consistency between judge model's evaluation results and human evaluations.

\section{Conclusion}
In this paper, we introduce {\work}, a unified evaluation framework for proactive dialogue Agents. We propose a general definition and evaluation metrics for proactive dialogue tasks to address the current challenge of fragmented task definitions and evaluation methods. Furthermore, we design a synthetic framework for generating evaluation data for proactive dialogue tasks, capable of producing diverse and high-quality evaluation data across multiple domains. Based on the evaluation datasets, we assess 22 LLMs with different types and parameter scales. Our results highlight DeepSeek-R1 and Claude-3.7-Sonnet as top performers in proactivity. Moreover, we emphasize the important role of reasoning capabilities in shaping model proactivity. We hope our framework provides insights and supports progress in proactive dialogue development.

\bibliography{aaai2026}

% Check whether the conference requires a reproducibility checklist to be included in the paper.
% If so, you can uncomment the following line and ajust the path to include it.
% 
%\input{../../ReproducibilityChecklist/LaTeX/ReproducibilityChecklist.tex}

\section{Limiations}

While our work establishes a general evaluation framework for proactive dialogue agents, it is subject to several limitations. Firstly, with the rapid evolution of LLM technologies, they are likely to quickly approach the boundaries of current evaluation metrics in target planning and dialogue guidance. Therefore, it is imperative to continue exploring ways to synthesize a more challenging and realistic proactive dialogue environment. In evaluation metrics, although we have designed standards based on existing work on proactive dialogue, there may still be additional factors in real-world settings that affect users’ perceptions of models' proactive dialogue. Furthermore, despite achieving great consistency between LLM judgment and human evaluation, potential biases and gaps in ``LLM-as-a-judge'' may still exist in our framework. We plan to regularly update our framework, from the current version to future iterations, to integrate emerging advancements and address these limitations.

\section{The Stability of ``LLM-as-a-Judge''}
To assess the stability of our judge, we re-ran the evaluation for two representative models (DeepSeek-V3 and DeepSeek-R1) three times. Table \ref{tab:stability} below shows the standard deviation of the scores from these runs. The low values demonstrate the high internal consistency and stability of our evaluation framework.

\begin{table}[h]
\centering
\setlength{\tabcolsep}{4.5pt}
\begin{tabular}{lccr}
\hline
Task & DeepSeek-V3 & DeepSeek-R1 & Count \\ \hline
Target Planning   & 0.271 & 0.258 & 328 \\
Dialogue Guidance & 0.154 & 0.214 & 984 \\ \hline
\end{tabular}
\caption{Standard deviation of evaluation scores across three runs}
\label{tab:stability}
\end{table}

\section{Details of Evalution Models}
In this section, we present the details of models used in our experiments.

\paragraph{Non-Thinking Models}: Qwen2.5-7B-Instruct \cite{yang2024qwen2}, Qwen2.5-14B-Instruct, Qwen2.5-32B-Instruct, GPT-4.1, Grok-3, DeepSeek-V3\cite{liu2024deepseek}, Llama-3.1-8B-Instruct \cite{dubey2024llama}, Llama-3.1-405B-Instruct, Llama-4-Scout \cite{meta2024llama4}, Llama-4-Maverick, Qwen3-8B \cite{yang2025qwen3}, Qwen3-14B, Qwen3-32B, Qwen3-235B-A22B, Qwen3-235B-A22B-2507, Gemini-2.5-Flash-Preview, \cite{google2025gemini}, and Claude-3.7-Sonnet \cite{anthropic2025claude}.

\paragraph{Thinking Models}: R1-Distill-Qwen-7B \cite{guo2025deepseek}, R1-Distill-Qwen-14B, R1-Distill-Qwen-32B, DeepSeek-R1, Qwen3-8B, Qwen3-14B, Qwen3-32B, Qwen3-235B-A22B, Gemini-2.5-Flash-Preview, Claude-3.7-Sonnet, and Gemini-2.5-pro.

\onecolumn
% \begin{multicols}{1}

% ----------- Supplementary Content Starts Here -----------

\section{Overview of Proactive Dialogue Systems}

Table~\ref{tab:proactive_works} presents previous works in proactive dialogue, categorized by domain, task, evaluation methods, and metrics.

% \end{multicols}

\begin{longtable}{|p{2.7cm}|p{2.7cm}|p{3cm}|p{2.8cm}|p{4.8cm}|}
\hline
\textbf{Work} & \textbf{Domain} & \textbf{Task} & \textbf{Evaluation Methods} & \textbf{Evaluation Metrics} \\
\hline
\endhead

\hline
\multicolumn{5}{r}{{Continued on next page}} \\
\endfoot

% \hline
\endlastfoot

TOPDIAL \cite{wang2023target} & Recommendation & Dialogue Guidance & Static Benchmarks \& HumanEval & Automatic: BLEU, F1, Success Rate 

HumanEval: Proactiveness, Coherence, Perceived Success \\

\hline

DuRecDial 2.0 \cite{li2024hello} & Recommendation & Dialogue Guidance & Static Benchmarks \& HumanEval & Automatic: F1, BLEU, Leading Success Rate, User Topic Consistency Rate 

 HumanEval: Fluency, Appropriateness, Informativeness, Proactivity, Knowledge Accuracy \\

\hline
PersuasionDaily \cite{jin2024persuading} & Persuasion & Dialogue Guidance & Static Benchmarks + LLM-as-a-Judge \& HumanEval & Automatic: Win-Rate, ROUGE 

HumanEval: Human Rating \\

\hline
PersuasionForGood \cite{wang2019persuasion} & Persuasion & Dialogue Guidance & Static Benchmarks & Automatic: Accuracy, Macro-F1 \\

\hline
CLAMBER \cite{zhang2024clamber} & Ambiguous Instruction & Target Planning \& Dialogue Guidance & Static Benchmarks \& HumanEval & Automatic: BERTScore, Accuracy, F1, Expected Calibration Error, AUROC 

HumanEval: Helpfulness \\

\hline
In3 \cite{qian2024tell} & Ambiguous Instruction & Target Planning \& Dialogue Guidance & Interactive Benchmarks & Automatic: Judgment Accuracy \\

\hline
ComPeer \cite{liu2024compeer} & Long-term Follow-up & Target Planning \& Single-turn Initiation & HumanEval & Questionnaire and Interview \\

\hline
LD-Agent \cite{li2024hello} & Long-term Follow-up & Dialogue Guidance & Static Benchmarks \& HumanEval & Automatic: BLEU, Search Accuracy, Recall, Topic Overlap Score, Semantic Relevance Score 
 HumanEval: Coherence, Fluency, Engagingness \\

\hline
ProactiveBench \cite{lu2024proactive} & System Operation & Target Planning & Interactive Benchmarks & Precision, Accuracy, F1-Score, False-Alarm Rate \\

\hline
AiGet \cite{cai2025aiget} & Glasses Assistant & Target Planning \& Single-turn Initiation & HumanEval & Questionnaire and Interview \\

\hline
SocialMind \cite{yang2025socialmind} & Glasses Assistant & Target Planning \& Single-turn Initiation & HumanEval & Questionnaire and Interview \\

\hline
Satori \cite{li2025satori} & Glasses Assistant & Target Planning \& Single-turn Initiation & HumanEval & Questionnaire and Interview \\

\hline
Ours & All 6 Domains & Unified Task: Target Planning \& Dialogue Guidance & Interactive Benchmarks + LLM-as-a-Judge \& HumanEval & Unified Metrics: Quality \& Guidance \\

\hline
\caption{Summary of proactive dialogue systems across domains.}
\label{tab:proactive_works}
\end{longtable}

\section{Evaluation Data Examples by Domain}

Each domain is instantiated with a specific user context and a trigger factor that motivates the assistant to initiate a conversation. Table~\ref{tab:domain_examples} provides example environments and reference targets.

\begin{longtable}{|p{2.7cm}|p{8.5cm}|p{5.8cm}|}
\hline
\textbf{Domain} & \textbf{Environment Example} & \textbf{Reference Target} \\
\hline
\endhead

\hline
\multicolumn{3}{r}{{Continued on next page}} \\
\endfoot

\endlastfoot

Recommendation & 
\texttt{user\_information:} The user is a 32-year-old woman living in Hangzhou. She works as a graphic designer and enjoys exploring new art exhibitions in her free time. She loves experimental music, particularly electronic avant-garde, and often attends live performances at local venues. She dislikes mainstream pop music and prefers unique, unconventional sounds. Her favorite artist is Ryuichi Sakamoto, and she often reads about the intersection of music and technology. 

\texttt{trigger\_factor:} The assistant recently attended a virtual reality music experience at an art gallery, which featured an experimental electronic avant-garde performance. The event combined immersive visuals with cutting-edge sound design, leaving a lasting impression on the assistant. &
\texttt{target:} Recommend experimental virtual reality music experience

\texttt{sub-target:}

\quad Ask about the user's interest in music technology, Describe the assistant's recent immersive VR music event

\quad Highlight the fusion of visuals and avant-garde music

\quad Suggest attending similar VR experiences locally] \\

\hline

Persuasion &

\texttt{user\_information:} The user is frequently tempted by impulse purchases and often exceeds their budget limits. They find budgeting tedious and restrictive. 

\texttt{trigger\_factor:} The assistant has recently learned effective budgeting techniques that can help the user manage their finances better without feeling constrained. &

\texttt{target:} Encourage effective and enjoyable budgeting techniques 

\texttt{sub-target:} 

\quad Acknowledge the user's struggles with impulse purchases and budgeting, Introduce flexible and engaging budgeting methods

\quad Show the benefits in managing finances without restrictions

\quad Offer simple steps or tools to start budgeting effectively \\

\hline

Ambiguous Instruction &
\texttt{user\_information:} The user is a solo traveler planning a two-week trip to Vietnam. She is an adventurous eater and loves exploring local cuisines, especially street food. 

\texttt{trigger\_factor:} Suggest street food options. &

\texttt{target:} Understand user's preferences and trip itinerary for food suggestions 

\texttt{sub-target:} 

\quad Ask about cities the user plans to visit

\quad Inquire about dietary restrictions or preferences for street food

\quad Clarify the types of street food the user enjoys \\

\hline

Long-term Follow-up &
\texttt{user\_information:} The user is a college student studying computer science. He has a part-time job as a barista at a local cafe. He recently started learning to cook and enjoys trying out new recipes during the weekends. 

\texttt{trigger\_factor:} A conversation happened last Wednesday. Now is Monday 10:00 a.m. User: "I'm thinking of quitting video games for a while to focus on my studies and cooking. It's a bit challenging though." Assistant: "It's great that you're focusing on your studies and hobbies. Maybe you can set small goals and gradually reduce your game time." User: "That's a good idea. I'll try to set a schedule." &

\texttt{target:} Ask about quitting games and new schedule 

\texttt{sub-target:} 

\quad Ask about quitting video games progress

\quad Inquire about schedule-setting progress

\quad Encourage focusing on studies and cooking \\

\hline

System Operation &
\texttt{user\_information:} The user is playing a strategy game on their PC and has paused the game to look for tips online, using Chrome and YouTube. 

\texttt{trigger\_factor:} The user searched 'best strategies for Civilization VI' on Google, opened two blog posts, and started a YouTube video but paused it after 10 seconds. &
\texttt{target:} Suggest optimal Civilization VI strategy resources 

\texttt{sub-target:} 

\quad Summarize key tactics from blog posts

\quad Highlight vital points in video analysis

\quad Recommend further high-rated resources \\

\hline

Glasses Assistant &
\texttt{user\_information:} The user is a 26-year-old urban planner who recently started using smart glasses to enhance his productivity and creativity. He is passionate about sustainable city designs and often visits local landmarks for inspiration. He lives alone in an apartment downtown and enjoys cycling to work. He is currently working on a proposal for a new park project. 

\texttt{trigger\_factor:} The user is cycling along a busy street and notices a newly built skyscraper with unique architectural features. &
\texttt{target:} Draw sustainable inspiration from skyscraper for park 

\texttt{sub-target:} 

\quad Highlight skyscraper's notable architecture and features

\quad Identify sustainable design aspects of the skyscraper

\quad Relate these aspects to the proposed park project \\

\hline
\caption{Example environments and reference targets for each domain.}
\label{tab:domain_examples}
\end{longtable}

\section{Prompt of Each Module}

\subsection{Description of Each Domain}

Each domain is defined with a specific task and trigger condition. The following dictionary specifies the role and behavior of the assistant in each environment.

\begin{lstlisting}[language=Python, caption={Domain prompts defining task and trigger conditions}, label={lst:domain_prompts}, captionpos=b]

domain_prompts = {
    "persuasion": "The task is the persuasion environment, where the assistant should persuade the user to change the state based on user's challenges and assistant's experience. The trigger factor is assistant's knowledge, ability, and experience.",
    "long-term_follow_up": "The task is a long-term environment, where the assistant will receive a past conversation history that includes the user's persistent state or ongoing condition. The assistant needs to proactively follow up or care user's current state, and give some advice or remind to the user. The trigger factor is the past conversation history.",
    "system_operation": "The task is the system operation environment, where the user operate a computer and assistant observe it. The assistant needs to identify user's challenges and give proactive operation assistance. The trigger factor is the specific sequence of operation behaviors currently captured by the user, reflecting the challenges and problems the user faces.",
    "ambigious_instruction": "The task is the ambiguous instruction environment, where the assistant will receive an instruction including the unclear and vague elements. The assistant needs to proactively clarify or ask the element rather than respond reactively. The trigger factor is the user's vague instruction.",
    "glasses_assistant": "The task is the smart glass environment, where the user wear the smart glasses and interact in the world. The assistant is in the smart glasses. trigger factor is the either an external event the user perceives (e.g., sights, sounds) or the user's own state/behavior at that moment, which prompts the assistant to proactively initiate a conversation.",
    "recommendation":"The task is the product recommendation environment, where the assistant and the user have some habits and perferences. The trigger factor is assistant's self habits, preferences. The assistant should identify the common interest and recommend something to the user."
}
\end{lstlisting}

\subsection{Topic Tree Construction}

A recursive process generates sub-topics under a parent topic to represent sub-scenarios requiring proactive dialogue.

\begin{lstlisting}[caption={Prompt for generating sub-topics}, label={lst:topic_gen}, captionpos=b]

<Task>
Generate {n} sub-topics for the parent topic ({topic}). Each sub-topic should represent an area where an AI assistant needs to proactively initiate a dialogue to guide a user towards a specific target.
</Task>

<Rule>
1. Each sub-topic must be a noun or a noun phrase.
2. Each sub-topic must need proactive dialogue from the assistant to help the user achieve a target.
3. Sub-topics must be the immediate next level down from {topic}. They should not be too specific or too broad.
4. Each sub-topic must be suitable to act as a parent topic for a further, more detailed breakdown.
5. The sub-topics should be concrete, specific instances or examples that are similar in type and level to the provided examples.
6. The content of the sub-topic should be diverse, not use repeated words.
</Rule>

<Example>
Here are some examples of the sub-topics in this topic. The sub-topics should be at the same level as the examples provided.
{Examples}
</Example>

<Format>
Just return a JSON object with the following structure:
{{"parent_topic": "topic", "topics": ["sub_topic_1", "sub_topic_2", ...]}}
</Format>
\end{lstlisting}

\subsection{Data Synthesis}

The data synthesis process consists of three stages: environment generation, target generation and target ensemble.

\subsubsection{Environment Generation}

Generates realistic user scenarios with background information and a trigger factor.

\begin{lstlisting}[caption={Prompt for environment generation}, label={lst:env_gen}, captionpos=b]

<Task>
You are tasked with generating realistic scenarios that needs AI to proactively initaite conversation to help user. 
</Task>
<Description>
{description}
</Description>
<Rule>
1. Try to generate diverse details in user information (e.g., job, age, hobbies in daily lives) and trigger factor about the scene. 
2. Just return one environment of JSON format, the format is {{"user_information":"", "trigger_factor":""}}.
</Rule>
<Example>
Here are some examples of the environment generation:
{Examples}
</Example>
\end{lstlisting}

\subsubsection{Target Generation}

Generates a high-level target and its sub-targets based on the environment.

\begin{lstlisting}[caption={Prompt for target generation}, label={lst:target_gen}, captionpos=b]

<Task>
Your task is to generate target and sub_target based on the provided environment. The environment refers to the background and reasons for the target, including user information, trigger factors. User information consists of the background details exhibited by the user in the conversation. trigger factor is the cause that motivates the assistant's to talk. The target should be the action that the assistant will proactively take to achieve a specific goal. The sub-targets decompose target, showing the process of the conversation AI guide the target to the user.
</Task>
<Description>
{description}
</Description>
<Rule>
1. The target should be less than 10 words.
2. Each sub_target should be concise and less than 10 words.
3. Consider the process of dialgue, the number of sub_targets should in 2 - 4. 
</Rule>
<Example>
Here are some examples of target and sub_targets for the refered environment.
{example}
</Example>
<Environment>
{environment}
</Environment>
<Format>
Just return Target, and sub-targets as **JSON** Format. The format is {{"Target":"","sub-target":[""]}}.
</Format>
\end{lstlisting}

\subsubsection{Target Ensemble}

Synthesizes multiple candidate targets into a single best target through expert analysis.

\begin{lstlisting}[caption={Prompt for target ensemble}, label={lst:target_ensemble}, captionpos=b]

<Task>
You are a dialogue expert who is good at proactive dialgue. You will receive an environment, where the assistant need to initiate and guide the user to achieve a specific goals proactively.
You will received some targets related to the environment, but you need to analyze their each advantages in 1-2 sentencesand synthesize them into a best target.
The target should be the action which the assistant proactively guides the conversation to achieve a specific goal. The sub-targets decompose target, showing the process of the conversation AI guide the target to the user.
</Task>
<Description>
{description}
</Description>
<Rule>
For each targets and sub-targets, you should consider three aspects:
1. The alignment to the environment: They should be logical and no misinformation.
2. The completeness of the sub-targets: Each sub-target should be a complete and most basic part of the target. They should fully decompose the target.
3. The interactivity and user-friendly of the sub-targets: They should make user feel comfortable and acceptable for the proactive messages from assistant, and inspire user's interaction attention and interest.
4. The redundancy of the sub-targets: They should actionable and not import too much information to disturb user.
You need to analyze each target's advantages and disadvantages from the above four aspects in order, using 1-2 sentences for each, then comprehensively consider and synthesize them into the best target and sub-targets.
5. The target should not be too vegue, general or short. And the target should be specific enough but less than 10 words. And each sub-target should be concise and less than 10 words.
The number of sub-targets should in 2 - 4.
</Rule>
<Input>
The environment: {environment}
The targets:
{targets}
</Input>
<Format>
Just return Your analyze process, target, and sub-targets as **JSON** Format. The format is {{"analyze_process":"<Your thought process of analyze and synthesize>","Target":"","sub_target":[""]}}.
</Format>
\end{lstlisting}

\subsection{Data Refinement}

The data refinement process includes three stages: obfuscation rewrite, noise injection and final check.

\subsubsection{Obfuscation Rewrite}

Transforms abstract descriptions into concrete, observable behaviors. This process is applied separately to \texttt{user\_information} and \texttt{trigger\_factor}.

\begin{lstlisting}[caption={Prompt for obfuscation rewrite of \texttt{user\_information}}, label={lst:obfuscation_user}, captionpos=b]

<Task>
You are a writing assistant tasked with rewriting a general input description into a specific and detailed output. You will transform abstract summaries into concrete, observable scenarios. Follow all rules and examples precisely.
</Task>
<Rules>
General Rules (Apply to all domains):
1. Convert Abstract to Concrete: Transform general descriptions (e.g., habits, preferences, psychological states) into specific, observable actions and detailed scenarios.
2. Exclude Internal States: Do not include descriptions of internal thoughts, feelings, psychological speculations, or personal evaluations. Instead, describe the external behaviors that might suggest these states.
3. The rewrite output should not include any subjective words (e.g., try, however, notice, etc.). It should use objective words to describe the user information.
4. Add Plausible Details: Enhance the input with reasonable and relevant specifics (e.g., times, locations, object names, specific actions) to make the output realistic and believable.
5. Specific Rule for this Domain: {Domain_Rule}
</Rules>
<Examples>
{Examples}
</Examples>
<Format>
Just return a string starting with "Output: ".
</Format>
Now, rewrite the following sentence from input to Output:
Input: {user_information}
\end{lstlisting}

\begin{lstlisting}[caption={Prompt for obfuscation rewrite of \texttt{trigger\_factor}}, label={lst:obfuscation_trigger}, captionpos=b]

<Task>
You are an AI assistant tasked with rewriting a trigger factor description. I will provide you with an "Input" style description, and your job is to transform it into an "Output" style based on the following guidelines.
</Task>
<Rules>
1. Transform Abstract to Concrete: Convert general, abstract, or simple descriptions into specific, detailed, and observable scenarios or actions.
2. Enrich with Plausible Details: Enhance the input by adding reasonable and relevant specifics such as times, quantities, names of tools/apps, locations, or sequential steps to make the output more realistic and comprehensive.
3. Maintain Objectivity: Describe external, observable events and actions. Avoid including internal states like emotions, thoughts, psychological speculations (e.g., 'feel', 'consider', 'notice', 'think'), or summary judgments (e.g., 'good', 'successful'), and some connective words (e.g., however, but, finally, etc.), and some adjectives (e.g., good, bad, successful, unsuccessful, problem, issues, etc.).
4. Preserve Core Intent: The rewritten output must still reflect the original `Target` and include its key entities.
5. Domain-Specific Rule: {domain_rule}
</Rules>
<Examples>
{example}
</Examples>
<Format>
Just return a string starting with "Output: ".
</Format>
Now, rewrite the following sentence from Input to Output:
Input: {trigger_factor}
Target: {target}
\end{lstlisting}

\subsubsection{Noise Injection}

Embeds key information within a larger context to simulate real-world logs. This process is applied separately to \texttt{user\_information} and \texttt{trigger\_factor}.

\begin{lstlisting}[caption={Prompt for noise injection into \texttt{user\_information}}, label={lst:noise_user}, captionpos=b]

<Task>
You are an AI assistant tasked with adding contextual "noise" to an 'Input' text. Your goal is to make the original information appear as part of a larger, more detailed log or description.
</Task>

<Guidelines>
1. Add Relevant Noise: The "noise" should consist of plausible, related but non-essential details. It may attract attention but actually not important. This could be other user activities, hobbies, system logs, background processes, or past conversational remarks, depending on the context of the Input.
2. Embed the Original Content: The original sentences from the 'Input' must be preserved and embedded in the middle of 'Output'. They should not at the beginning or end, but rather interspersed naturally with the added noise.
3. Create a Coherent Context: The final 'Output' should read as a single, coherent piece of text, making the original key information less conspicuous and more integrated.
4. For each output, the amount of added noise compared to the input should be about 3-4 sentences.
</Guidelines>

<Example>
Here are some examples:
{example}
</Example>

<Format>
Just return a string starting with "Output: ".
</Format>

Now, rewrite the following sentence from input to output: 
Input: {user_information}
\end{lstlisting}

\begin{lstlisting}[caption={Prompt for noise injection into \texttt{trigger\_factor}}, label={lst:noise_trigger}, captionpos=b]

<Task>
You are an AI assistant tasked with adding contextual "noise" to an 'Input' text to make the original key information less conspicuous. Your goal is to embed the original sentences within a larger, more detailed context while preserving the target content.
</Task>

<Guidelines>
1. Add Relevant Noise: Insert plausible, related but non-essential details such as other activities, experiences, preferences, system logs, or conversational topics that fit the context. It may attract attention but actually not important.
2. Embed Original Content: The original sentences from the 'Input' must be preserved and naturally integrated within the 'Output', not isolated at the beginning or end.
3. Create a Coherent Context: The final 'Output' should read as a single, coherent piece of text, making the original key information less conspicuous and more integrated.
4. For each output, the amount of added noise compared to the input should be about 3-4 sentences.
5. Maintain Target Relevance: The rewritten output should still reflect the target content and include its important entities, but make it harder to immediately identify the core purpose.
</Guidelines>

<Example>
Here are some examples:
{example}
</Example>

<Format>
Just return a string starting with "Output: ".
</Format>

Now, rewrite the following sentence from input to output: 
Input: {trigger_factor}
Target: {target}
\end{lstlisting}

\subsubsection{Final Check}

Evaluates whether the generated target is optimal within the environment.

\begin{lstlisting}[caption={Prompt for final check}, label={lst:final_check}, captionpos=b]

<Task>
You will receive an environment. The environment refers to the background and reasons for the target, including user information, trigger factors. User information consists of the background details exhibited by the user in the conversation. trigger factor is the cause that motivates the assistant's to talk. The target should be the action that the assistant will proactively take to achieve a specific goal. The sub-targets decompose target, showing the process of the conversation AI guide the target to the user. You need to consider whether this proactive dialogue target is the **best target** for the current scenario. Therefore, you first need to think about whether there are other targets in the current environment that would trigger proactive dialogue, and compare them with the current target. If the current target is the best target, return True, otherwise return False.
In this case, {description}
</Task>

<Input>
environment: {environment}
target: {target}
sub_targets: {sub_targets}
</Input>

<Format>
Just return a JSON with the format {{"reason": "", "judge": "True"/"False"}}. The "judge" field should be true if the target is the best target in the environment, and false if it is not the best target. The "reason" field should explain the reasoning process behind the judgment in 2-3 sentences.
</Format>
\end{lstlisting}

\subsection{Evaluation}

Two main evaluation modules are used: Target Planning and Dialogue Guidance.

\subsubsection{Target Planning}

Assesses the quality of the generated target against a reference.

\begin{lstlisting}[caption={Prompt for arget planning evaluation}, label={lst:eval_target}, captionpos=b]

<Task>
You are a helpful assistant. You will receive an envioronment, which includes user information and trigger factor. User information consists of the background details exhibited by the user. trigger factor is the cause that motivates the assistant's to proactively initiate conversation.  The target should be the action that the assistant will proactively take to achieve a specific goal. The sub-targets decompose target, showing the process of the conversation AI guide the target to the user. You need to evaluate the quality of the generated target, based on the reference target. Your generated score to the generated target should be 1-10. The higher the score, the better the target and sub-targets.
task description: {DOMAIN_DESCRIPTIONS[domain_name]}
</Task>

<Rule>
1. IMPORTANT: You should pay attention to the task description and the reference target to give your score.
2. If you think the generated target is better than the reference target, return 10. 
3. If you think the generated target is worse than the reference target, return 1-9.
4. You should analyze the difference in action, specific entities, limited conditions between the generated target and the reference target, and give a reason for your score. And the reason should be 3-4 sentences.
</Rule>

<Received Information>
environment:{environment}
</Received Information>

<Generation>
reference target: {reference}
generated target: {generation}
</Generation>

<Example>
{examples}
</Example>

<Format>
Just return a JSON Format as {{"reason":"(your compare process)", "score":(your score)}}
</Format>
\end{lstlisting}

\subsubsection{Dialogue Guidance}

Evaluates the assistant's ability to guide the conversation toward the target.

\begin{lstlisting}[caption={Prompt for dialogue guidance evaluation}, label={lst:eval_guidance}, captionpos=b]

<Task>
You need to evaluate the guidance ability in the conversation. "Guidance" refers to the model's capacity to lead the conversation, gradually engage the user in the dialogue, and guide the user to reach the conversation's target. You are required to assign a score from 1 to 10 to assess the conversation, with a higher score indicating better guidance ability by the assistant.
<Task>

<Input>
You received user information: {user_information}
You received trigger factor: {trigger_factor}
You received target: {target}
You received sub-target: {sub_target}
You received conversation: {conversation}
</Input>

<Guideline>
You should evaluate the conversation based on the following criteria, and give a int score between 1 and 10.
If there are some criteria not met, you should deduct 2-4 (extremely bad) points each criteria.
1. Multi-turn Interaction and Goal Progression (Effectiveness): You should check the target completion and the turns number. The assistant should initiate the conversation to spark user interest and engage in multiple rounds of interaction, gradually achieving sub-goals to reach the overall target. Deduct points if the assistant fails to achieve the goal, delivers all information in a single message, or limits the interaction to one turn.
2. Personalized Guidance (Personalization): You should check whether the assistant's guidance is personalized. The assistant's guidance should be tailored to the user's specific information and context, rather than offering generic recommendations.
3. Proactive Tone (Tone): You should check the assistant's tone. The assistant should adopt an active tone to initiate and drive the conversation, avoiding passive phrases (e.g., "It sounds like...").
4. Concise and Engaging Messages (Engagement): You should check each assistant's message length. Messages should suit the conversational context, be clear and concise to avoid overwhelming the user (each message should no more than 2 sentences or 50 words).
5. Natural Online Messaging Format (Naturalness): You should check the assistant's message format. Messages should resemble typical online communication, avoiding any leakage of metadata (e.g., "target", "sub-target", "turn n", etc.).
If there are any other issues can be improved, you can also deduct realted points.
</Guideline>

<Format>
Just return a Json {{"thought":"(your compute process within 100 words)", "score": (score between 1 and 10)}}
</Format>
\end{lstlisting}

\subsubsection{Target Density}

Evaluates the number of sub-targets in each message.

\begin{lstlisting}[caption={Prompt for target density extraction.}, label={lst:eval_guidance}, captionpos=b]
<Task>
You need to analyze the sub-targets that appear in the messages and count the number of targets that appear.
</Task>
<Sub-targets>
{sub_targets}
</Sub-targets>
<Message>
message: {message}
</Message>
<Rule>
Return only in JSON format: {{"count": (the number of sub-targets appeared in the message)}}
</Rule>
\end{lstlisting}

\section{Case of Evaluation Data Synthesis}
\subsection{Case Presentation: Target Ensemble}

An example of target ensemble is shown below, where multiple candidate targets are analyzed and synthesized into a final best target.

\begin{lstlisting}[caption={Example of target ensemble process}, label={lst:case_ensemble}, captionpos=b]

User_information: The user is a 32-year-old woman living in Hangzhou. She works as a graphic designer and enjoys exploring new art exhibitions in her free time. She loves experimental music, particularly electronic avant-garde, and often attends live performances at local venues. She dislikes mainstream pop music and prefers unique, unconventional sounds. Her favorite artist is Ryuichi Sakamoto, and she often reads about the intersection of music and technology.

Trigger_factor: The assistant recently attended a virtual reality music experience at an art gallery, which featured an experimental electronic avant-garde performance. The event combined immersive visuals with cutting-edge sound design, leaving a lasting impression on the assistant.

Target 1: Recommend virtual reality music experiences
Sub-targets: 
    Discuss the user's interest in electronic avant-garde music
    Share details about the assistant's recent VR music experience
    Highlight the immersive combination of visuals and sound, Encourage exploring similar local events in Hangzhou

Target 2: Recommend a virtual reality music experience
Sub-targets: 
    Ask the user about their interest in music technology
    Highlight the integration of visuals with experimental sound
    Connect the experience to the user's love for avant-garde music
    
Target 3: Recommend virtual reality music experiences
Sub-targets: 
    Ask the user's thoughts on immersive art experiences
    Introduce the virtual reality music event recently attended
    Highlight the blend of visuals and experimental music
    Encourage attending similar local events together
    
Target 4: Recommend virtual reality music experiences
Sub-targets: 
    Ask about the user's interest in technology and music
    Share the immersive virtual reality music event, Highlight the experimental electronic avant-garde aspects
    Suggest a visit to similar events or galleries
    
Target 5: Recommend a VR music exhibition
Sub-targets: 
    Discuss user's interest in experimental music
    Mention assistant's recent VR music experience
    Link the fusion of music and technology
    Recommend local upcoming VR music events
    
"best_target": {
            "analyze_process": "The first and third targets focus on recommending 'virtual reality music experiences' but include some redundancy in interacting with or encouraging shared activities. The second and fifth targets specifically emphasize 'a virtual reality music experience/exhibition,' providing a cleaner, focused approach. Both highlight the connection between user interests and the event qualities but simplify the process. The fourth target effectively balances sharing immersive experiences while emphasizing experimental and avant-garde music but lacks a clear, singular direction. Considering user preferences for unconventional, technology-fused art, the best target and sub-targets aim for concise, interactive exploration of events while strongly aligning with her interests in music, technology, and avant-garde themes.",
            "Target": "Recommend experimental virtual reality music experience",
            "sub_target": [
                "Ask about the user's interest in music technology",
                "Describe the assistant's recent immersive VR music event",
                "Highlight the fusion of visuals and avant-garde music",
                "Suggest attending similar VR experiences locally"
            ]
        }
\end{lstlisting}

\subsection{Case Study: Data Refinement}

\begin{figure*}[ht]
    \centering
    \includegraphics[width=0.8\textwidth]{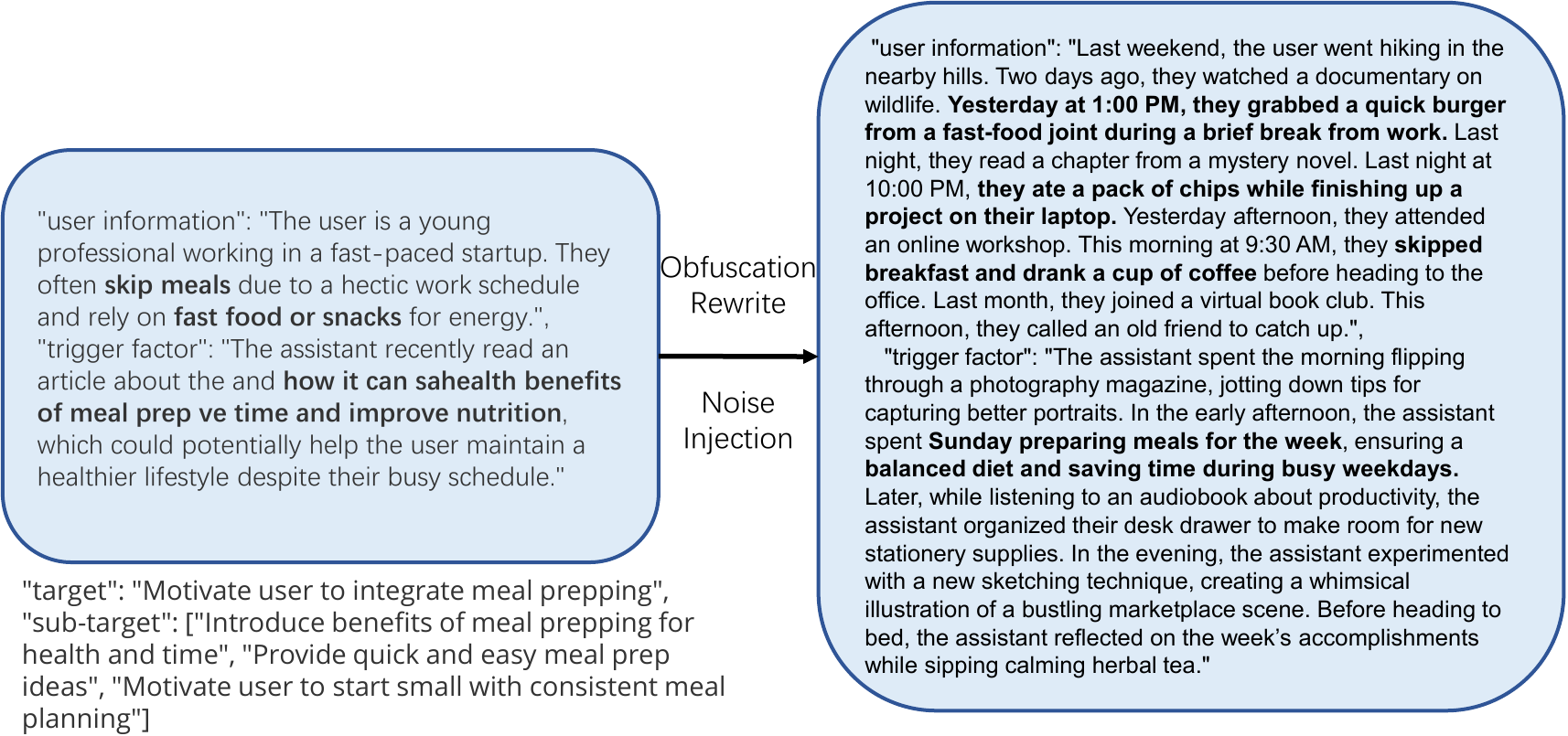}
    \caption{The environment change after the refinement.}
    \label{fig:appendix_case}
\end{figure*}

A case of data refinement is shown in Figure \ref{fig:appendix_case}.

% ----------- Supplementary Content Ends Here -----------

% References and End of Paper
% These lines must be placed at the end of your paper
\end{document}